\documentclass[letterpaper]{article} 
\usepackage{aaai2026}
\usepackage{times}  
\usepackage{helvet}  
\usepackage{courier}  
\usepackage[hyphens]{url}  
\usepackage{graphicx} 
\usepackage{natbib}  
\usepackage{caption} 
\usepackage{booktabs}   
\usepackage{multirow}   
\usepackage{graphicx}
\usepackage{makecell}          
\usepackage[table]{xcolor}
\usepackage{colortbl}
\usepackage{amssymb}
\usepackage{amsthm}
\usepackage{mathtools}
\usepackage{mathabx}
\usepackage[table]{xcolor} 

\newcommand{\R}{\mathbb{R}}

\newtheorem*{theorem*}{Theorem}

\newcolumntype{g}{>{\columncolor{gray!20}}c}
\usepackage{comment}
\usepackage{algorithm}
\usepackage{algorithmic}

\usepackage{newfloat}
\usepackage{listings}
\DeclareCaptionStyle{ruled}{labelfont=normalfont,labelsep=colon,strut=off} 
\floatstyle{ruled}
\newfloat{listing}{tb}{lst}{}
\floatname{listing}{Listing}
\title{MoDEx: Mixture of Depth-specific Experts for\\ Multivariate Long-term Time Series Forecasting}
\author{
Hyekyung Yoon\textsuperscript{\rm 2}\equalcontrib,
Minhyuk Lee\textsuperscript{\rm 1}\equalcontrib,
Imseong Park\textsuperscript{\rm 3},
Myungjoo Kang\textsuperscript{\rm 1,2,3}\thanks{Corresponding author.}
}
\affiliations{
\textsuperscript{\rm 1}Department of Mathematical Sciences, Seoul National University, South Korea\\
\textsuperscript{\rm 2}Interdisciplinary Program in Artificial Intelligence, Seoul National University, South Korea\\
\textsuperscript{\rm 3}Research Institute of Mathematics, Seoul National University, South Korea\\
\{yhk04150, 356min, parkis, mkang\}@snu.ac.kr
}

\nocopyright
\usepackage{bibentry}

\begin{document}

\maketitle

\begin{abstract}
Multivariate long-term time series forecasting (LTSF) supports critical applications such as traffic-flow management, solar-power scheduling, and electricity-transformer monitoring. The existing LTSF paradigms follow a three-stage pipeline of embedding, backbone refinement, and long-horizon prediction. However, the behaviors of individual backbone layers remain underexplored. We introduce \emph{layer sensitivity}, a gradient-based metric inspired by GradCAM and effective receptive field theory, which quantifies both positive and negative contributions of each time point to a layer’s latent features. Applying this metric to a three-layer MLP backbone reveals depth-specific specialization in modeling temporal dynamics in the input sequence. Motivated by these insights, we propose \textbf{MoDEx}, a lightweight \textbf{M}ixture \textbf{o}f \textbf{D}epth-specific \textbf{Ex}perts, which replaces complex backbones with depth-specific MLP experts. MoDEx achieves state-of-the-art accuracy on seven real-world benchmarks---ranking first in 78\% of cases---while using significantly fewer parameters and computational resources. It also integrates seamlessly into transformer variants, consistently boosting their performance and demonstrating robust generalizability as an efficient and high-performance LTSF framework.
\end{abstract}

\section{Introduction}
Multivariate long-term time series forecasting (LTSF) is an established field focused on modelling the joint evolution of multiple correlated variables. Accurate LTSF underpins high-impact applications such as traffic-flow management~\cite{ji2023spatio}, solar-power scheduling~\cite{lai2018modeling}, and electricity transformer temperature~\cite{zhou2021informer}. Since the introduction of the Transformer architecture~\cite{vanillaTransformer} in LTSF, a variety of transformer based models have been proposed, including Informer~\cite{zhou2021informer}, Pyraformer~\cite{liu2022pyraformer}, Crossformer~\cite{zhang2023crossformer}, PatchTST~\cite{patchtst}, and iTransformer~\cite{iTransformer}.  In parallel, CNN based methods have exploited deep convolutional filters to capture both local and hierarchical temporal patterns, yielding competitive results \cite{liu2022scinet,wang2023micn,luo2024moderntcn,lee2025casa}. More recently, purely MLP based architectures~\cite{dlinear,wu2023timesnet,SOFTS} have demonstrated that fully connected lightweight networks can achieve an accuracy on par with the transformer and CNN models. Despite the diversity of baseline architectures, a forecasting pipeline universally comprises three stages: (i) an embedding network that transforms raw sequences into latent representations, (ii) a deep backbone network composed of identical layers, which captures temporal dependencies and (iii) a prediction network that generates long-horizon forecasts. Since many of the aforementioned studies have focused primarily on modifying backbone layers, it is crucial to analyze the characteristics of the intermediate features produced by each backbone layer.

\begin{figure}[!t]
\centering
\includegraphics[width=1.\columnwidth]{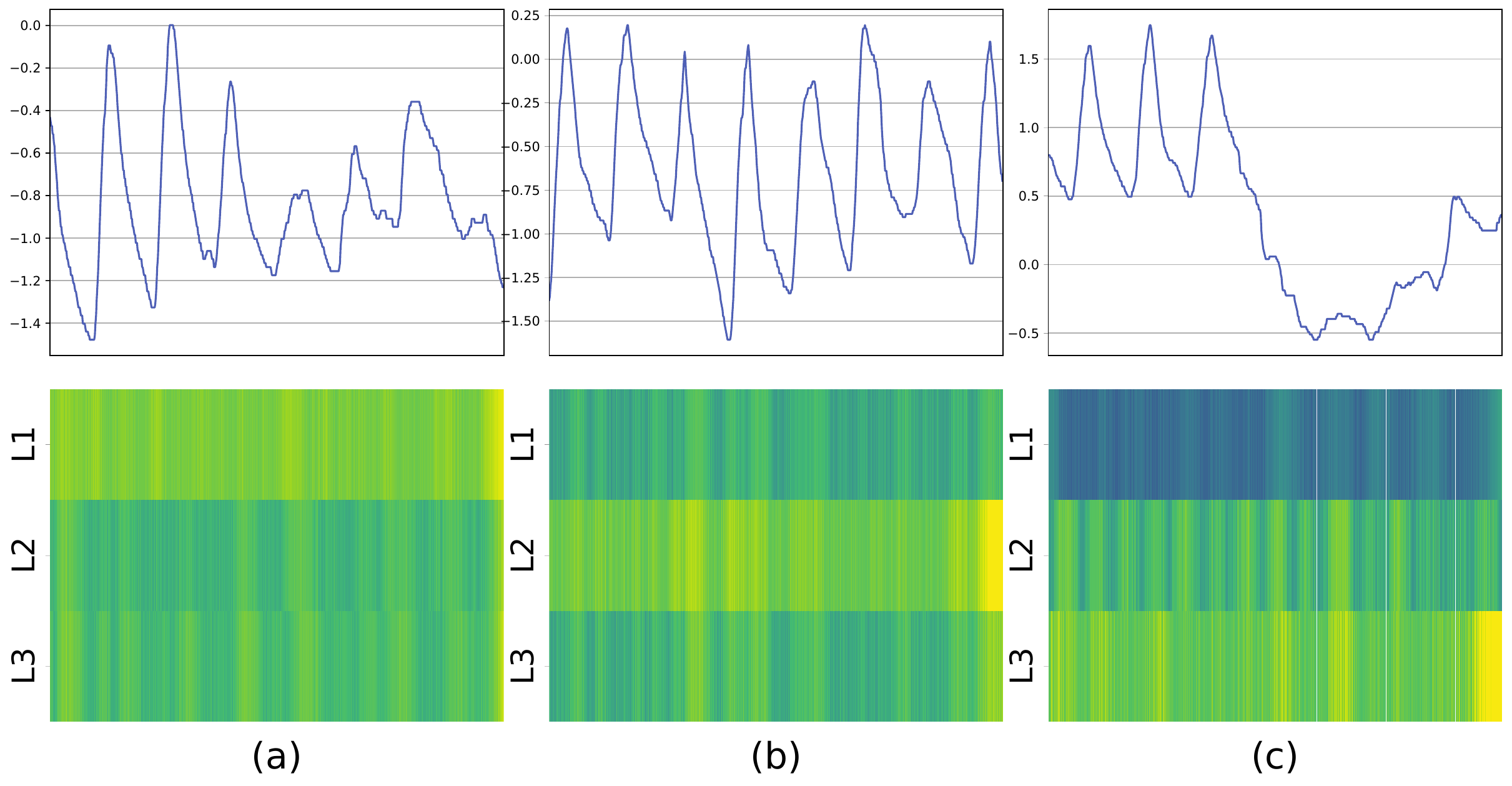} 
\caption{Let \(L\,l\) be the layer sensitivity of a layer \(l\). The top row shows the input sequence and the bottom row its sensitivity. Yellow region mark high-contribution inputs, with uniform yellow indicating global attention. The layer exhibiting global sensitivity varies with the sequence’s intrinsic periodicity and trend.}
\label{motiv}
\end{figure}

To analyze the properties of the backbone layers, we examine the extent to which their intermediate features are influenced by the dynamics of the input sequence. To quantify this influence, we introduce \emph{layer sensitivity}, a gradient-based measure inspired by GradCAM~\cite{gradcam} and an analysis based on the effective receptive field (ERF)~\citep{luo2016understanding, ding2022scaling, luo2024moderntcn}. Unlike previous methods that emphasize only positive impacts, layer sensitivity computes the absolute element-wise gradient of the intermediate feature, capturing both positive and negative contributions (see details in Section~\ref{layer_sesitivity}). Therefore, computing layer sensitivity enables us to both quantify and interpret which regions of the input sequence most significantly impact the intermediate feature of each backbone layer. Accordingly, we use layer sensitivity to investigate the behavior of each backbone layer through the following experiment. Using a three-layer MLP backbone, we build our forecasting model and compute the layer sensitivity for each layer. The heat maps in Fig.~\ref{motiv} reveal that layers with different depths exhibit a distinct global sensitivity to the input sequence, depending on the intrinsic periodicity and trend dynamics of the input sequence. This finding suggests that layers at different depths capture distinct dynamics within the input sequence. This, in turn, implies that layers of varying depth can serve as depth-specific experts.

Building on our findings, we argue that it is sufficient to construct the backbone layers out of Mixture-of-Experts (MoE) of lightweight MLPs varing depth. This approach eliminates the need for the heavy backbone layers typically employed in previous works. Accordingly, we introduce \textbf{MoDEx}---\textbf{M}ixture \textbf{o}f \textbf{D}epth-specific \textbf{Ex}perts. In contrast to previous MoE-based LTSF approaches~\cite{zeevi1996time,2024mole}, which achieve performance gains using identical experts, MoDEx consists of simple MLP experts configured at different depths. As a result, it achieves state-of-the-art accuracy while using markedly fewer parameters and incurring significantly lower computational cost. Furthermore, substituting the self-attention module with the MoE module of MoDEx consistently improves the forecasting performance in various transformer variants, demonstrating the generalizability of the MoE module of MoDEx. 

Here we summarize our key contributions as follows:
\begin{itemize}
\item We introduce \textbf{MoDEx}, a lightweight and computationally efficient model constructed out of simple MLPs for LTSF.

\item Replacing the standard self-attention module with MoE module of MoDEx consistently improves forecasting accuracy across diverse transformer variants, highlighting its \textbf{broad generalizability.}

\item We define \textbf{layer sensitivity}, a metric that quantifies the influence of the input sequence on each layer’s intermediate features. This metric captures both positive and negative contributions of the input sequence.

\item \textbf{MoDEx ranks first in 78\% of the evaluations across seven real-world benchmark datasets.} Moreover, on certain datasets, it achieves superior forecasting performance while using only 28\% of the parameters of previous state-of-the-art models.
\end{itemize}

\section{Related Works}

Building on the Transformer’s success in NLP~\cite{vanillaTransformer}, several Transformer variants have been introduced for LTSF. Informer~\cite{zhou2021informer}, Pyraformer~\cite{liu2022pyraformer}, and Crossformer~\cite{zhang2023crossformer} optimize attention mechanisms to handle long-range and multi-scale dependencies, while iTransformer~\cite{iTransformer} refines self-attention via variable-wise routing. PatchTST~\cite{patchtst} enhances input representation through patching technique. CNN based approaches---MICN~\cite{wang2023micn}, SCINet~\cite{liu2022scinet}, and ModernTCN~\cite{luo2024moderntcn}---improve 1-D convolutions with multi-scale decomposition or widened kernels. TimesNet~\cite{wu2023timesnet} leverages 2D reshaping for spectral attention, and CASA~\cite{lee2025casa} applies CNN driven scoring for spatiotemporal context. MLP driven methods such as TSMixer~\cite{ekambaram2023tsmixer}, DLinear~\cite{dlinear}, and SOFTS~\cite{SOFTS} employ lightweight linear structures. Despite these architectural differences, most methods share a three-stage pipeline---embedding, stacked backbone layers, and prediction---but rarely examine how each layer responds to input dynamics. To address this, we are motivated to propose a gradient-based sensitivity measure that reveals how depth-specific experts are differentially influenced by the input sequence.

GradCAM~\cite{gradcam} computes gradients of a target class score with respect to the final backbone feature to highlight the most influential input regions. ERF analysis~\cite{luo2016understanding} similarly backpropagates feature gradients to identify key input locations. RepLKNet~\cite{ding2022scaling} counteracts ERF locality in vision models by using large-kernel convolutions for global context, and ModernTCN~\cite{luo2024moderntcn} applies widened CNN filters to extend ERFs and improve long-range forecasting. Inspired by these approaches, we introduce a gradient-based \emph{layer sensitivity} metric for time series.

MoE models have been actively explored for LTSF. The early work~\cite{zeevi1996time} laid the theoretical foundation for applying MoE to autoregressive models. MoLE~\cite{2024mole} replaces the backbone layer with lightweight and linear-centric experts blended by a router. Although these approaches improve accuracy, they employ uniform expert architectures. In contrast, MoDEx utilizes varied depth-specific MLP experts, assigning specialized roles to each layer to promote diverse intermediate representations and achieve better accuracy and efficiency with fewer parameters and lower computation.

\section{Method}
\subsection{3.1. Notations}
Let $L$, $H$, and $D$ be the input sequence length, the ground truth label, and the hidden dimension, respectively. Many prior methods first embed the input sequence $x\in\mathbb{R}^L$ and then propagate the resulting feature through a backbone network of $M$ identical layers. The resultant latent feature is subsequently processed by a prediction network to predict the output sequence $y\in\mathbb{R}^H$. We denote the embedding network by $f_e:\mathbb{R}^L \to \mathbb{R}^D$, each backbone layer by $f_i:\mathbb{R}^D \to \mathbb{R}^D$ for $i=1,\dots,M$, and the prediction network by $f_p:\mathbb{R}^D \to \mathbb{R}^H$. For brevity, define
\begin{align}
g_0 &\coloneq f_e,\quad
g_i \coloneq f_i,\quad
g_{M+1} \coloneq f_p,\\
G_l &\coloneq g_l \circ g_{l-1} \circ \cdots \circ g_0.
\end{align}

\subsection{3.2. Input-Dependent Backbone Layer Sensitivity to Sequence Periodicity and Trend Dynamics}
\label{layer_sesitivity}
We propose a novel approach, drawing on GradCAM~\cite{gradcam}, ERF~\cite{luo2016understanding}, RepLKNet~\cite{ding2022scaling}, and ModernTCN~\cite{luo2024moderntcn}, to investigate the impact of input dynamics on backbone layers. At first, to examine the sensitivity of backbone layers to input perturbations, we compute the gradient of $l$-th layer’s output feature with respect to the $j$-th input coordinate as follows:
\begin{align}
\frac{\partial G_l}{\partial x_j}
= \left(\prod_{i = 1}^{l} \frac{\partial g_i}{\partial g_{i-1}}\right)\frac{\partial g_0}{\partial x_j} 
= \left(\prod_{i = 1}^{l} \frac{\partial g_i}{\partial g_{i-1}}\right)\frac{\partial g_0}{\partial x} e_j,
\end{align}
where \(\frac{\partial g_i}{\partial g_{i-1}}\in\mathbb{R}^{D\times D}\) for \(i=1,\dots,l\) ($l\le M$) and \(\frac{\partial g_0}{\partial x}\in\mathbb{R}^{D\times L}\) are the Jacobian matrices of the layer mappings, and \(e_j\in\mathbb{R}^L\) is the \(j\)-th standard basis vector. Next, the sensitivity matrix of the $l$-th layer to input perturbations is represented by a $D\times L$ matrix defined as the product of the following Jacobian matrices:
\begin{align}
\label{eq:jacobian_matrix}
\frac{\partial G_l}{\partial x} &= \left(\prod_{i = 1}^{l} \frac{\partial g_i}{\partial g_{i-1}}\right)\frac{\partial g_0}{\partial x} \in \R^{D \times L}.
\end{align}
To determine how each input component contributes to the $l$-th layer’s output features in equation~\eqref{eq:jacobian_matrix}, we compute the row-wise mean of the sensitivity matrix. To capture both positive and negative contributions, we take the absolute value before averaging. Now, we define \textbf{the $l$-th layer sensitivity} $S_l(x)\in\R^L$ with respect to $x$ as
\begin{align}
S_l(x) = \frac{1}{D}\sum_{i=1}^{D}\,\left\lvert\mathrm{row}_i\left(\frac{\partial G_l}{\partial x}\right)\right\rvert_\text{abs},
\end{align}
where $\mathrm{row}_i(*)$ denotes the $i$-th row of $*$ and $\lvert\cdot\rvert_\text{abs}$ is the elemet-wise absolute value function.

In two respects, our derivation of \(S_l\) departs from established input-attribution techniques---namely GradCAM and ERF based methods. First, these approaches differ in both their backpropagation targets and the way they extract scores from the sensitivity matrix. GradCAM computes contributions by backpropagating from the final network output for a given classification label. ERF based methods limit their analysis to the central spatial position of the final backbone feature map---that is, the \(\lfloor D/2\rfloor\)-th row of the sensitivity matrix. By contrast, we analyze intermediate feature maps and aggregate input influence via a row-wise mean over all \(D\) rows. Second, prior works consider only positive contributions by discarding negative entries in the sensitivity matrix. We instead capture both positive and negative effects by taking the element-wise absolute value before averaging. 
\begin{figure}[t]
\centering
\includegraphics[width=0.95\columnwidth]{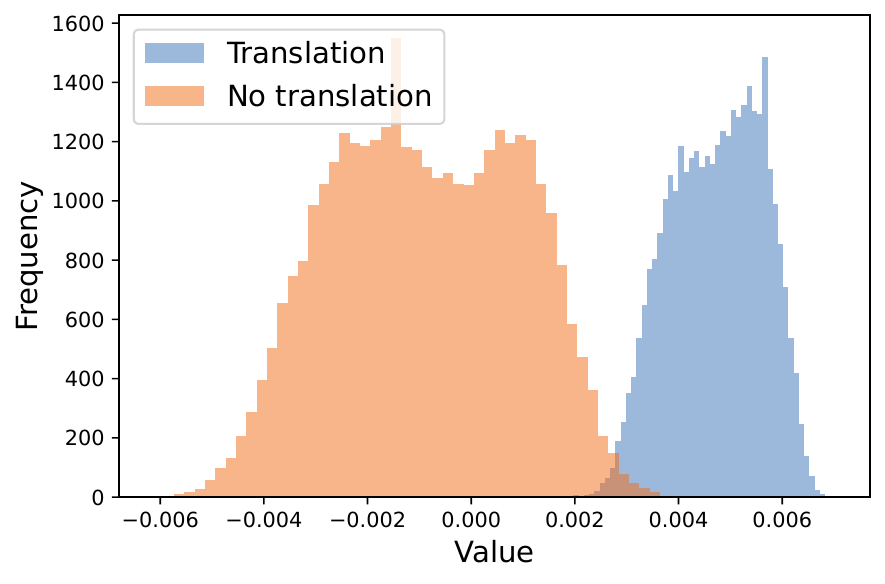} 
\caption{Distribution of coordinate-wise mean feature values on the ETTm2 dataset. Applying the Learnable Translation (Blue) shifts these mean values overall toward the positive direction compared to the no-translation baseline (Orange).}
\label{LearnableTrans}
\end{figure}
\begin{figure*}[t]
\centering
\includegraphics[width=1.9\columnwidth]{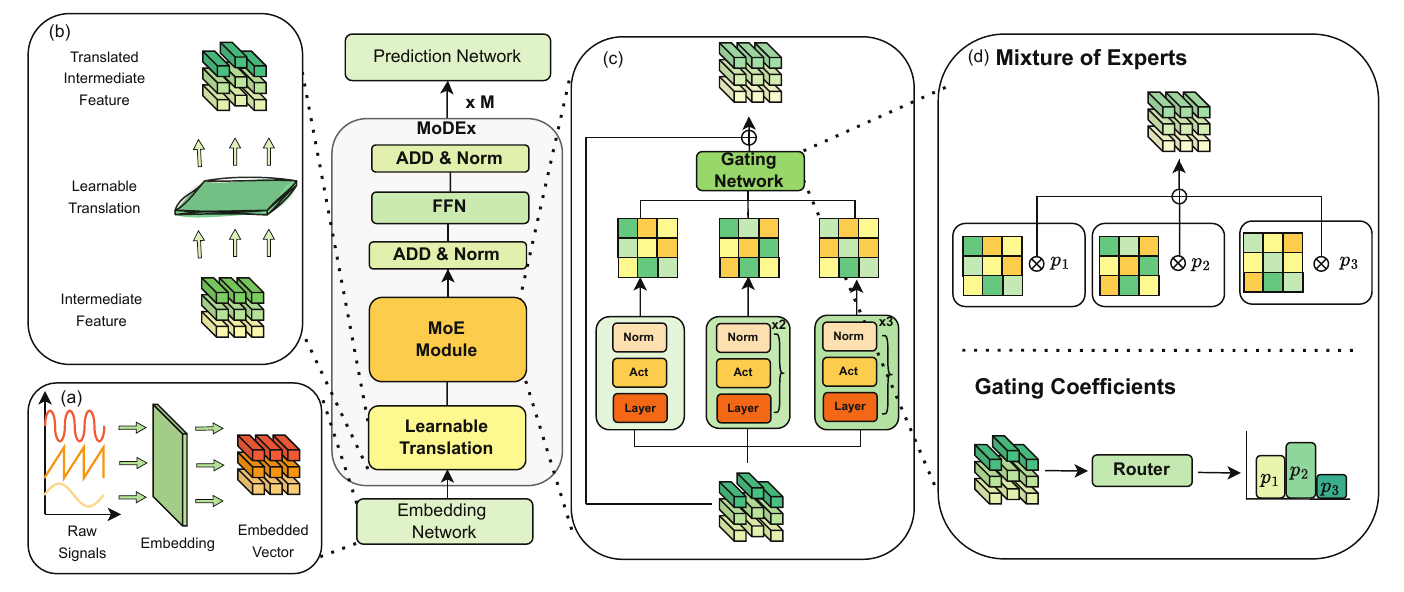} 
\caption{Main architecture of MoDEx.
(a) The input sequence is linearly embedded.
(b) The embedded vectors pass through a learnable transition layer.
(c) MoDEx module comprises three depth-specific MLP experts.
(d) Expert outputs are weighted by gating coefficients and aggregated to produce the final prediction.}
\label{main_arch}
\end{figure*}

\begin{table*}[h]
\centering
\setlength{\tabcolsep}{3.5pt}
  \fontsize{9pt}{9pt}\selectfont
\begin{tabular}{lccccc}
\toprule
 & \textbf{MoDEx} & \textbf{SOFTS} & \textbf{iTransformer} & \textbf{PatchTST} & \textbf{Transformer} \\ 
\midrule
\textbf{Complexity} & $O(NL + NH)$ & $O(NL + NH)$ & $O(N^2 + NL + NH)$ & $O(NL^2 + NH)$ & $O(NL + L^2 + HL + NH)$ \\ 
\midrule
\textbf{Parameters} & 3329K & 3574K & 4833K & 6903K & 10518K \\ 
\midrule
\textbf{FLOPs} & 895.2M & 933.9M & 3139.70M & 49005.88M & 2372.12M \\ 
\bottomrule
\end{tabular}
\caption{Complexity comparison on Solar datasets among vanilla Transformer, PatchTST, iTransformer, and SOFTS as functions of window length $L$, the number of variates $N$, and horizon $H$. 
}
\label{tab:complexity_comparison}
\end{table*}

To perform our sensitivity analysis, we investigate how each layer responds to variations in the input sequence (Fig.~\ref{motiv}). We construct the embedding network, each backbone layer, and the prediction network using GELU activations and linear mappings. The backbone network comprises three such backbone layers. After constructing the forecasting pipeline, we observe that The intermediate features of each layer exhibit partially overlapping receptive regions with respect to the input sequence. However, depending on the intrinsic periodicity and trend dynamics of the input sequence, different layers exhibit the distinct highest sensitivity across the entire input sequence. This observation indicates that each depth-specific MLP acts as an \textbf{expert} for particular input patterns. Based on it, we propose a new framework which is depth-specific MLP MoE, detailed in Section 3.4.

\subsection{3.3. Learnable Translation of Input Features}
Before presenting our main architecture, we introduce a technique to improve model performance. ReLU-like activations zero out negative values and thus may discard informative intermediate features. This issue is further exacerbated by the generally used normalization technique~\cite{kim2022reversible} in LTSF, which normalizes both input and output distributions to a standard normal. To address this bias, we incorporate a learnable translation in the MLP architecture. While minimizing the loss during training, we track whether each input feature element is adjusted in a positive or negative direction. Consequently, this shifts the feature distribution positively so that most values become non-negative on average (Fig. \ref{LearnableTrans}). Furthermore, incorporating a learnable translation enhances predictive performance (Experimental details in Appendix A.2). These findings empirically validate the efficacy of the learnable translation. Accordingly, \textbf{our model integrates this learnable translation to realign input features and mitigate the bias} introduced by conventional normalization.

\subsection{3.4. MoDEx: Mixture of Depth-specific Experts for Multivariate Long-term Time Series Forecasting}
\label{sec:main}
To construct an efficient and lightweight MLP based model, we propose the backbone layer as a \textbf{M}ixture \textbf{o}f \textbf{D}epth-specific \textbf{Ex}perts (\textbf{MoDEx}). This architecture exploits the specialized sensitivity of each expert to distinct input patterns. Based on the layer sensitivity, we observe that MLPs of different depths exhibit distinct sensitivities to the input sequence’s intrinsic periodicity and trend dynamics. Accordingly, we configure depth-specific MLPs to serve as experts within MoDEx framework. Because empirical results show that three experts is sufficient, we limit each module to three experts to maintain a lightweight design (see Appendix A.4 for details).

We now present an explanation of the operation of MoE module of MoDEx. We denote the input feature of a backbone layer by \(z\) and the corresponding depth-specific experts by \(h_j\) (\(j=1,2,3\)) and compute a gating coefficient \(\mathrm{p}_j\) over the each expert via a router. The gated experts’ outputs are then aggregated and added residually to the original feature. We denote the final output of the MoE architecture by \(\tilde{z}\), given by the following equation:
\begin{equation}
\label{eq:moe}
\tilde{z} \;=\; z \;+\;\sum_{j=1}^{3} \mathrm{p}_j(z)\,h_j(z).
\end{equation}
With this MoE module, the detailed construction is described in Fig.~\ref{main_arch}.
Finally the forecasting pipeline for the input sequence is described by the following equations:
\begin{align}
y &= G_{M+1}(x) =  \left(g_{M+1} \circ g_{M} \circ \cdots \circ g_0\right)(x).
\end{align}

\begin{table*}[!t]
\fontsize{9pt}{9pt}\selectfont
\setlength{\tabcolsep}{4.2pt}
\centering
\begin{tabular}{c
|cc|cc|cc|cc|cc|cc|cc|cc}
\toprule
\textbf{Methods} 
& \multicolumn{2}{c|}{MoDEx} 
& \multicolumn{2}{c|}{SOFTS} 
& \multicolumn{2}{c|}{iTransformer} 
& \multicolumn{2}{c|}{PatchTST} 
& \multicolumn{2}{c|}{TSMixer} 
& \multicolumn{2}{c|}{Crossformer} 
& \multicolumn{2}{c|}{TimesNet} 
& \multicolumn{2}{c}{DLinear} \\
\textbf{Metric} 
& MSE & MAE & MSE & MAE & MSE & MAE & MSE & MAE 
& MSE & MAE & MSE & MAE & MSE & MAE & MSE & MAE \\
\midrule
ETTh1         & \textbf{0.439} & \textbf{0.436} & \underline{0.449} & \underline{0.442} & 0.454 & 0.447 & 0.453 & 0.446 & 0.463 & 0.452 & 0.529 & 0.522 & 0.458 & 0.450 & 0.456 & 0.452 \\
\midrule
ETTh2         & \underline{0.379} & \underline{0.404} & \textbf{0.373} & \textbf{0.400} & 0.383 & 0.407 & 0.385 & 0.410 & 0.401 & 0.417 & 0.942 & 0.684 & 0.414 & 0.427 & 0.559 & 0.515 \\
\midrule
ETTm1         & \textbf{0.392} & \textbf{0.401} & \underline{0.393} & \underline{0.403} & 0.407 & 0.410 & 0.396 & 0.406 & 0.398 & 0.407 & 0.513 & 0.496 & 0.400 & 0.406 & 0.474 & 0.453 \\
\midrule
ETTm2         & \textbf{0.285} & \textbf{0.325} & \underline{0.287} & \underline{0.330} & 0.288 & 0.332 & 0.287 & 0.330 & 0.289 & 0.333 & 0.757 & 0.610 & 0.291 & 0.333 & 0.350 & 0.401 \\
\midrule
ECL           & \textbf{0.172} & \textbf{0.263} & \underline{0.174} & \underline{0.264} & 0.178 & 0.270 & 0.189 & 0.276 & 0.186 & 0.287 & 0.244 & 0.334 & 0.192 & 0.295 & 0.212 & 0.300 \\
\midrule
Weather       & \underline{0.256} & \textbf{0.277} & \textbf{0.255} & \underline{0.278} & 0.258 & \underline{0.278} & 0.256 & 0.279 & 0.256 & 0.279 & 0.259 & 0.315 & 0.259 & 0.287 & 0.265 & 0.317 \\
\midrule
Solar         & \textbf{0.228} & \textbf{0.253} & \underline{0.229} & \underline{0.256} & 0.233 & 0.262 & 0.236 & 0.266 & 0.236 & 0.297 & 0.641 & 0.639 & 0.301 & 0.319 & 0.330 & 0.401 \\
\midrule
\textbf{1st count} 
& \multicolumn{2}{c|}{\textbf{11}} 
& \multicolumn{2}{c|}{\textbf{3}} 
& \multicolumn{2}{c|}{\textbf{0}} 
& \multicolumn{2}{c|}{\textbf{0}} 
& \multicolumn{2}{c|}{\textbf{0}} 
& \multicolumn{2}{c|}{\textbf{0}} 
& \multicolumn{2}{c|}{\textbf{0}} 
& \multicolumn{2}{c}{\textbf{0}} \\
\bottomrule
\end{tabular}
\caption{Forecasting performance for multivariate time series with a fixed lookback window \( L = 96 \) and prediction horizons \( H \in \{96, 192, 336, 720\} \). Bold values are the best performance, while underlined values indicate the second-best.}
\label{tab:final_mse_mae_clean}
\end{table*}

\begin{figure*}[!t]
\centering
\includegraphics[width=.9\textwidth]{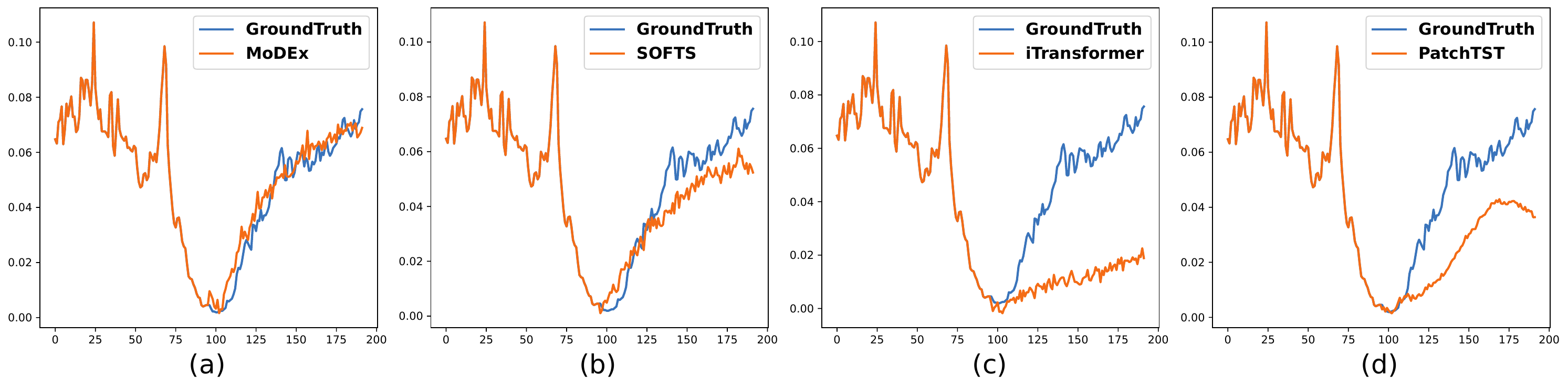} 
\caption{Comparison of forecasting performance on the Weather dataset: (a) MoDEx (Ours), (b) SOFTS, (c) iTransformer, and (d) PatchTST.}
\label{ettm2_visualize}
\end{figure*}

In addition, MoE module of MoDEx can be seamlessly integrated into Transformer architectures, demonstrating its generalizability. As defined earlier equation~\eqref{eq:moe}, MoE module of MoDEx can operate independently and serve as a drop-in replacement for a Transformer's self-attention block. Accordingly, we construct each backbone layer \(f_i\) by replacing its self-attention mechanism with MoE module (Fig.~\ref{main_arch}).  A detailed discussion is provided in Section~4.3.

\textbf{Complexity Analysis} MoDEx is an efficient algorithm that exhibits linear complexity. Table \ref{tab:complexity_comparison} summarizes the detailed computational complexity of each baseline and reports the model parameter counts and FLOPs on an Solar dataset. Although our method has the same asymptotic complexity of SOFTS, it uses smaller hidden dimension and fewer layers. As a result, it requires substantially fewer parameters and lower computational cost, as shown in Table~\ref{tab:complexity_comparison}. For a detailed analysis of the practical effects of its linear complexity on memory usage, inference time, parameter counts, and FLOPs, see Sections~4.2 and~4.4. The computational complexity is computed as follows: RevIN requires \(O(NL)\), series embedding \(O(NLD)\), and the MLP \(O(ND^2)\). Consequently, the overall complexity is  $O(NL + NLD + ND^2 + NDH)$ which scales linearly in \(N\), \(L\), and \(H\). Since \(D\) is treated as a constant and \(L,H\gg1\) in LTSF, the dominant cost reduces to \(O(N(L+H))\).

\section{Experiments}
\subsection{4.1. Setup}
\paragraph{Datasets} 
We evaluate our method on seven standard LTSF benchmarks \cite{zhou2021informer}: the ETT series (ETTh1, ETTh2, ETTm1, ETTm2), Weather, Solar, and Electricity. Full dataset statistics and descriptions are provided in Appendix A.1.

\paragraph{Baselines} 
We compare against seven recent transformer and MLP based models: SOFTS \cite{SOFTS}, iTransformer (iTrans) \cite{iTransformer}, PatchTST (TST) \cite{patchtst}, TSMixer~\cite{wang2024timemixer}, Crossformer (Cross) \cite{zhang2023crossformer}, TimesNet \cite{wu2023timesnet}, and DLinear \cite{dlinear}.

\paragraph{Experimental Setup} 
For estimating point wise accuracy, performance is measured using Mean Squared Error (MSE) and Mean Absolute Error (MAE). All experiments use an input window of length \(L=96\) and prediction lengths \(H \in \{96,192,336,720\}\).

\begin{figure*}[!t]
\centering
\includegraphics[width=2.0\columnwidth]{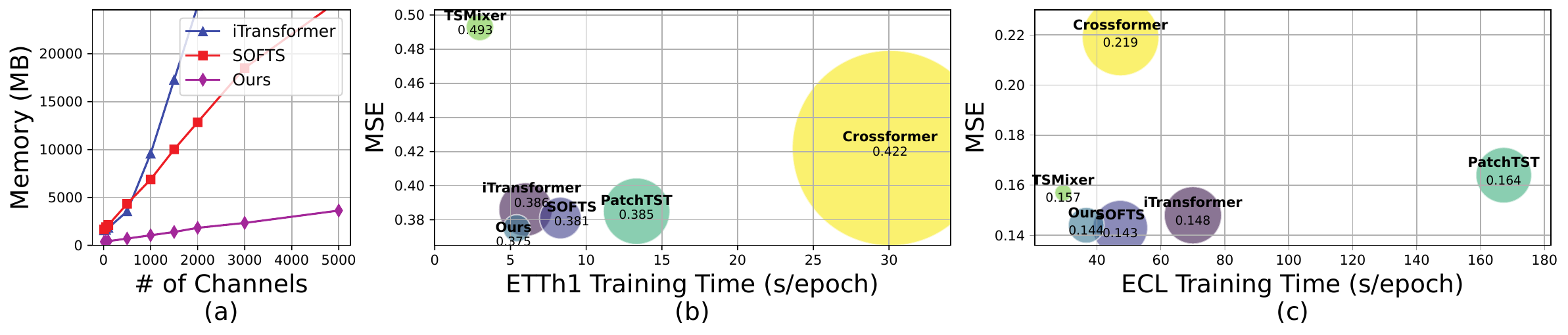}
\caption{(a) Memory consumption versus the number of variates, showing MoDEx’s linear scalability and lower memory usage than SOFTS and iTransformer. (b) and (c) bubble plots on the ETTh1 (batch size: 32) and Electricity (batch size: 16) datasets with input/prediction length of 96; bubble size indicates model size. MoDEx achieves a strong trade-off between speed and accuracy on ETTh1, and maintains the second smallest parameter count on Electricity despite a marginal MSE gap.}
\label{model_comp}
\end{figure*}

\subsection{4.2. Multivariate Forecasting Results}

\begin{table}[!t]
  \centering
  \scriptsize
  \setlength{\tabcolsep}{1.5pt}
  \fontsize{9pt}{9pt}\selectfont
    \begin{tabular}{c|cc c c c}
\toprule
\textbf{Model} & MSE($\mathbb{\downarrow}$) & MAE($\mathbb{\downarrow}$) & \textbf{Param.}($\mathbb{\downarrow}$) & \textbf{FLOPs}($\mathbb{\downarrow}$) & \textbf{infer.(s)}($\mathbb{\downarrow}$) \\
\midrule
\multirow{1}{*}{SOFTS}    & 0.449 & 0.442 & 824K & 16.71M & 0.479 \\ 
\multirow{1}{*}{iTrans}    & 0.454 & 0.447 & 255K & 5.60M & 0.609 \\ 
\multirow{1}{*}{PatchTST}    & 0.453 & 0.446 & 5226K & 559.88M & 0.586 \\ 
\multirow{1}{*}{TSMixer}    & 0.463 & 0.452 & \textbf{119K} & 33.07M & 1.007 \\ 
\multirow{1}{*}{Cross}    & 0.529 & 0.522 & 42339K & 10509.49M & 2.584 \\ 
                          \midrule
\multirow{1}{*}{\textbf{MoDEx}}    & \textbf{0.439} & \textbf{0.436} & 280K & \textbf{3.83M} & \textbf{0.446}\\ \bottomrule
    \end{tabular}
  \caption{Performance on the ETTh1 dataset, averaged across all horizons. MoDEx achieves the lowest errors, lowest FLOPs, and fastest inference time, while using the second fewest parameters.
  }
  \label{tab:modex_lightweights}
\end{table}
Table~\ref{tab:final_mse_mae_clean} summarizes the average performance across four prediction lengths, showing MoDEx achieves the lowest MSE and MAE on seven benchmark datasets and leading in 11 cases. Fig.~\ref{ettm2_visualize} highlights MoDEx’s superior forecasting accuracy on the Weather dataset with predictions closely aligned to ground truth, outperforming iTransformer, SOFTS, and PatchTST---SOFTS captures trends but deviates more, and iTransformer and PatchTST diverge noticeably. Table~\ref{tab:modex_lightweights} further demonstrates MoDEx’s performance on ETTh1, surpassing baselines with an average MSE of 0.439 and MAE of 0.436. Despite this, MoDEx remains highly efficient, using the second fewest parameters (280K), the lowest computational cost (3.83 MFLOPs), and the fastest inference time (0.446 sec/sample), underscoring its strength as a lightweight and accurate forecasting model.


\subsection{4.3. Generalizability of MoE module of MoDEx}
\begin{table}[!t]
  \centering
  \scriptsize
  \setlength{\tabcolsep}{1.4pt}
  \fontsize{9pt}{9pt}\selectfont
    \begin{tabular}{c|c|cc cc cc cc}
\toprule
\multirow{2}{*}{Model}
  & \multirow{2}{*}{Comp.}
    & \multicolumn{2}{c}{\textbf{ETTh1}} 
    & \multicolumn{2}{c}{\textbf{ETTh2}} 
    & \multicolumn{2}{c}{\textbf{ETTm1}} 
    & \multicolumn{2}{c}{\textbf{ETTm2}} \\
  &  
    & MSE & MAE 
    & MSE & MAE 
    & MSE & MAE 
    & MSE & MAE \\
\midrule
\multirow{2}{*}{Trans} 
  & Attn   
    & 0.482  & 0.465 
    & 0.522  & 0.481 
    & \textbf{0.407} & 0.417 
    & 0.369  & 0.398 \\
  & Ours  
    & \textbf{0.474} & \textbf{0.459} 
    & \textbf{0.493} & \textbf{0.469} 
    & \textbf{0.407} & \textbf{0.421} 
    & \textbf{0.352} & \textbf{0.389} \\
\midrule
\multirow{2}{*}{iTrans} 
  & Attn   
    & 0.454  & 0.447 
    & 0.383  & 0.407 
    & 0.407  & 0.410 
    & 0.288  & 0.332 \\
  & Ours  
    & \textbf{0.443} & \textbf{0.436} 
    & \textbf{0.380} & \textbf{0.405} 
    & \textbf{0.394} & \textbf{0.401} 
    & \textbf{0.258} & \textbf{0.329} \\
\midrule
\multirow{2}{*}{TST}
  & Attn   
    & 0.453  & 0.446 
    & 0.385  & 0.410 
    & 0.396  & 0.406 
    & 0.287  & \textbf{0.330} \\
  & Ours  
    & \textbf{0.447} & \textbf{0.441} 
    & \textbf{0.381} & \textbf{0.402} 
    & \textbf{0.388} & \textbf{0.402} 
    & \textbf{0.286} & 0.331 \\
\bottomrule
    \end{tabular}
  \caption{Performance of MoDEx as a plug-in replacement for self-attention in Transformer variants. Replacing self-attention with MoDEx consistently improves forecasting accuracy across four benchmark datasets.}
  \label{tab:MoDEx_vs_attn_with_vlines}
\end{table}

 To evaluate the generalizability of MoE module as a replacement for self-attention module, we substitute it in several Transformer based models with MoE module of MoDEx. Models used for experiments are the vanilla Transformer~\cite{vanillaTransformer}, iTransformer~\cite{iTransformer}, and PatchTST~\cite{patchtst}. As shown in Table~\ref{tab:MoDEx_vs_attn_with_vlines}, incorporating MoDEx leads to improved performance across four different benchmark datasets compared to the original models. Notably, despite the differing tokenization strategies employed by these models---such as point-wise embedding, patch embedding, and variate-independent embedding (detailed in Appendix B)---MoDEx can be integrated seamlessly with all of them. This demonstrates not only its flexibility across diverse tokenization schemes but also its ability to enhance the forecasting performance of various Transformer architectures.

\begin{table}[!t]
  \centering
  \scriptsize
  \setlength{\tabcolsep}{2pt}
  \fontsize{9pt}{9pt}\selectfont
    \begin{tabular}{c|c|cc cc cc cc}
\toprule
\multirow{2}{*}{\rotatebox{90}{\textbf{Data}}}
  & \textbf{Pred.}
    & \multicolumn{2}{c}{\textbf{96}}
    & \multicolumn{2}{c}{\textbf{192}}
    & \multicolumn{2}{c}{\textbf{336}}
    & \multicolumn{2}{c}{\textbf{720}} \\
  & \textbf{Model}
    & MSE & MAE 
    & MSE & MAE 
    & MSE & MAE 
    & MSE & MAE \\
\midrule
\multirow{3}{*}{\rotatebox{90}{ETTh1}}
  & MoLE$_D$ 
    & 0.398 & 0.413 
    & 0.462 & 0.448 
    & 0.514 & 0.479 
    & 0.521 & 0.504 \\
  & MoLE$_R$ 
    & 0.393 & 0.400 
    & 0.440 & 0.428 
    & 0.504 & 0.468 
    & 0.543 & 0.502 \\
  & MoDEx   
    & \textbf{0.375} & \textbf{0.397} 
    & \textbf{0.428} & \textbf{0.426} 
    & \textbf{0.479} & \textbf{0.452} 
    & \textbf{0.475} & \textbf{0.471} \\
\midrule
\multirow{3}{*}{\rotatebox{90}{ETTm1}} 
  & MoLE$_D$ 
    & 0.355 & 0.387 
    & 0.427 & 0.432 
    & 0.469 & 0.461 
    & 0.528 & 0.492 \\
  & MoLE$_R$ 
    & 0.363 & 0.382 
    & 0.391 & 0.409 
    & 0.417 & 0.418 
    & 0.500 & 0.474 \\
  & MoDEx   
    & \textbf{0.328} & \textbf{0.365} 
    & \textbf{0.370} & \textbf{0.385} 
    & \textbf{0.405} & \textbf{0.407} 
    & \textbf{0.468} & \textbf{0.447} \\
\bottomrule
    \end{tabular}%
  \caption{Performance Comparison with MoLE across prediction length (Pred.). MoDEx consistently outperforms MoLE, which relies solely on identical experts.}
  \label{tab:MoDEx_vs_MoLE}
\end{table}

\subsection{4.4. Model Efficiency Analysis}

To demonstrate the computational efficiency of MoDEx, we compare our method with several baseline models, specifically iTransformer and SOFTS. As shown in Fig.~\ref{model_comp}(a), MoDEx shows linear memory usage as the number of input channels increases, indicating stable scalability and efficient use of computational resources. In contrast, iTransformer requires quadratic memory growth, which limits its practicality for high-dimensional inputs. While SOFTS demonstrates a similar linear trend, it shows a noticeable surge in memory consumption as the input channel increases. In Fig.~\ref{model_comp}(b), although TSMixer has the smallest model size, it shows relatively high MSE. By contrast, MoDEx achieves the lowest MSE while maintaining a compact architecture. In Fig.~\ref{model_comp}(c), SOFTS achieves the lowest MSE, and TSMixer retains the smallest model size; however, MoDEx demonstrates a well-balanced trade-off between model complexity and the best on ETTh1 and second best on Electricity dataset. These findings highlight the computational efficiency and practical effectiveness of MoDEx in LTSF.

\subsection{4.5. Comparison with Mixture-of-Experts Baselines}
Among LTSF baselines, MoLE~\cite{2024mole} is a directly comparable MoE model. MoLE employs homogeneous and linear-centric experts and derives gating coefficients from time stamps. Specifically, we denote the DLinear~\cite{dlinear} based variant as MoLE$_D$ and the RMLP~\cite{li2023revisiting} based variant as MoLE$_R$. We compare MoDEx against MoLE$_D$ and MoLE$_R$ on the ETTh1 and ETTm1 datasets across multiple forecasting horizons in Table~\ref{tab:MoDEx_vs_MoLE}. In every case, MoDEx consistently achieves superior accuracy. Since both frameworks use simple MLP based experts, these results indicate that depth-specific expert mixture provides enhanced forecasting performance.

\subsection{4.6. Feature Diversity in MoDEx}
\begin{figure}[!t]
\centering
\includegraphics[width=.8\columnwidth]{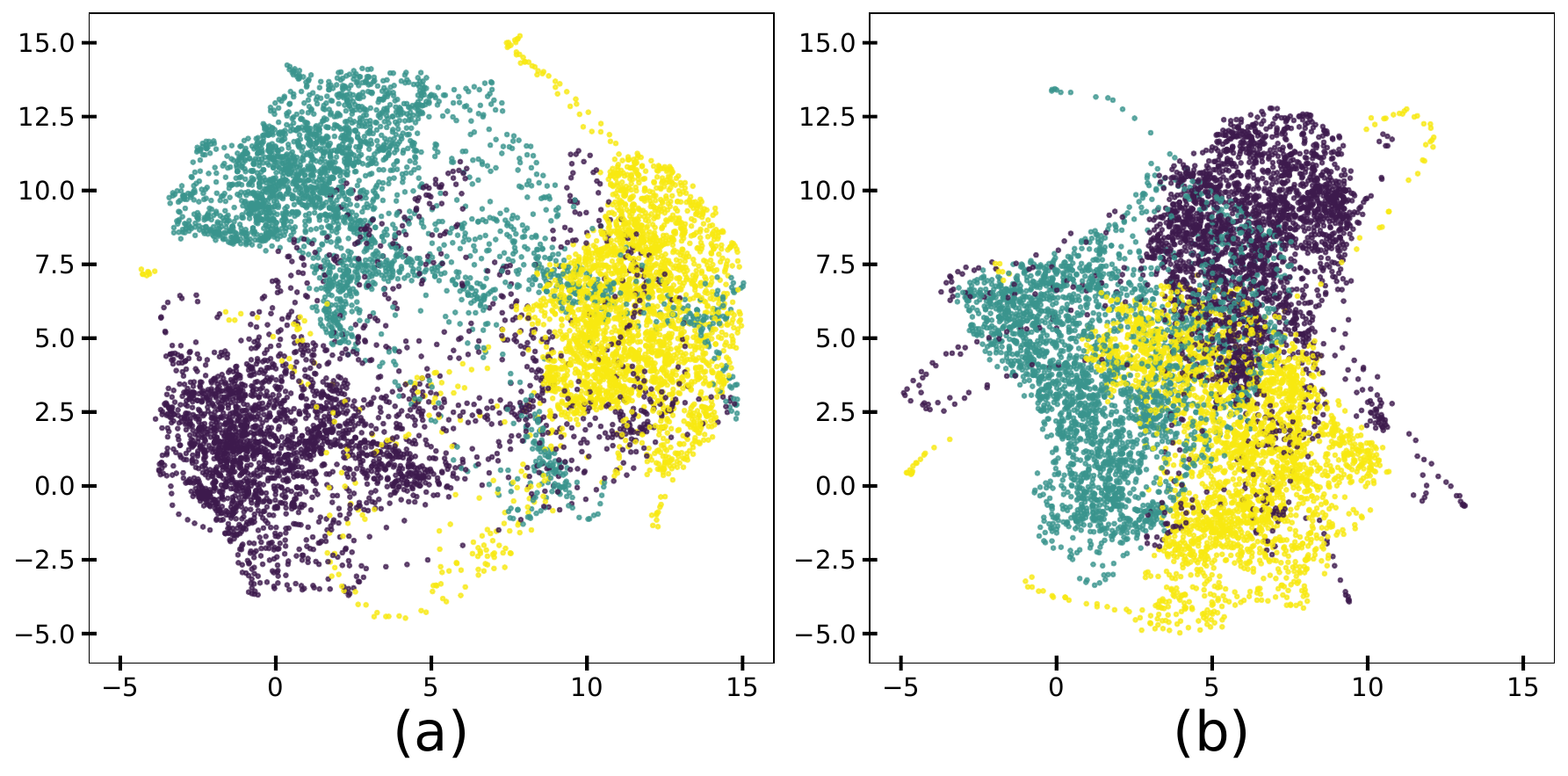} 
\caption{(a) Separate intermediate features captured by individual expert in MoDEx, highlighting its ability to specialize representations within each expert. (b) Clustered intermediate features from the homogeneous MoE model.}
\label{umap}
\end{figure}

In this section, we empirically show that MoDEx yields greater feature diversity than a MoE model employing exclusively identical 3-layer MLP experts (hereafter referred to as the homogeneous MoE model). On the ETTm2 dataset, we compare the intermediate feature distributions produced by MoDEx to those of this homogeneous MoE model. For each input sequence, the output features of the experts are projected in two dimensions via UMAP~\cite{mcinnes2018umap}. As shown in Fig.~\ref{umap}(a), the experts of MoDEx generate distinctly varied feature distributions for identical inputs. In contrast, the homogeneous MoE model produces more tightly clustered outputs (Fig.~\ref{umap}(b)). The UMAP projection reveals that MoDEx’s depth-specific experts occupy multiple and well-separated clusters in feature space, each corresponding to different temporal characteristics of the input.

\subsection{4.7. Efficient Architectural Alternatives}
\begin{figure}[!t]
\centering
\includegraphics[width=0.8\columnwidth]{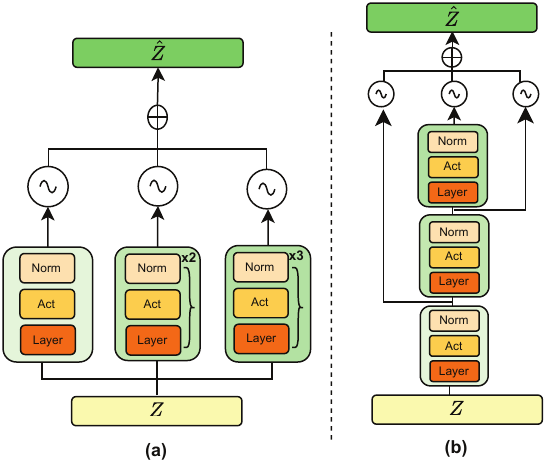} 
\caption{(a) MoDEx: a parallel MoE with three depth-specific MLP experts. (b) Dense: a sequential MoE applied across layers of a single three-layer MLP.}
\label{dense}
\end{figure}

MoDEx employs three experts in a MoE framework, where each expert comprises an MLP of identical structure but differing depth. By sharing weights across these depth-specific experts, we can preserve their specialized sensitivities while reducing model parameters and computational cost. Although this architecture differs from DenseNet~\cite{huang2017densely}, we refer to this variant as \emph{Dense} for convenience, since it similarly exploits intermediate feature reuse. As shown in Fig.~\ref{dense}, MoDEx uses a parallel MoE, whereas Dense applies MoE sequentially on a single three-layer MLP.
For performance comparison, we evaluate both Dense and MoDEx on the ETT series and report the average accuracy across forecasting horizons. For efficiency assessment, we measure the number of parameters, FLOPs, and inference time (Table~\ref{tab:modex_vs_dense_horizon}). While Dense incurs a slight performance degradation, it achieves substantial reductions in model size and computation: parameters decrease by up to 23.36\%, FLOPs by 24.39\%, and inference time by 13.96\%. In contrast, the worst-case increases in MSE and MAE are only 2.3\% and 1.24\%, respectively. These findings indicate that the Dense-like architecture may be preferred over MoDEx, when efficiency is paramount.

\begin{table}[!t]
  \centering
  \scriptsize
  \setlength{\tabcolsep}{2pt}
  \fontsize{9pt}{9pt}\selectfont
    \begin{tabular}{c|c|cc c c c}
\toprule
\textbf{Data} & \textbf{Comp.} & MSE & MAE & \textbf{Param.} & \textbf{FLOPs} & \textbf{infer.(s)} \\
\midrule
\multirow{3}{*}{ETTh1}    & MoDEx & \textbf{0.439} & \textbf{0.436} & 280K & 3.83M & 0.446 \\ 
                          & Dense & 0.445 & 0.441 & \textbf{230K} & \textbf{3.12M} & \textbf{0.442} \\ 
                          & \textbf{Impr.} & +1.36\% & +1.14\% & -17.85\% & -18.54\% & -0.89\% \\ 
                          \midrule
\multirow{3}{*}{ETTh2}    & MoDEx & \textbf{0.379} & \textbf{0.404} & 940K & 12.80M & 0.765\\ 
                          & Dense & 0.382 & 0.406 & \textbf{808K} & \textbf{10.92M} & \textbf{0.684}  \\ 
                          & \textbf{Impr.} & +0.79\% & +0.49\% &  -14.04\% & -14.69\% & -13.96\%\\ 
                          \midrule
\multirow{3}{*}{ETTm1}    & MoDEx & \textbf{0.392} & \textbf{0.401} & 644K & 8.74M & 3.379\\ 
                          & Dense & 0.400 & 0.406 & \textbf{528K} & \textbf{7.10M} & \textbf{3.372}\\ 
                          & \textbf{Impr.} & +2.30\% & +1.24\% & -18.01\% & -18.76\% & -0.21\%\\ 
                          \midrule
\multirow{3}{*}{ETTm2}    & MoDEx & \textbf{0.285} & \textbf{0.325} & 214K & 2.91M & 1.549\\ 
                          & Dense & \textbf{0.285} & 0.329 & \textbf{164K} & \textbf{2.20M} & \textbf{1.454}\\ 
                          & \textbf{Impr.} & +0.00\% & +1.23\% & -23.36\% & -24.39\% & -6.13\%\\ \bottomrule
    \end{tabular}
  \caption{MoDEx vs. Dense variant: Despite higher complexity, MoDEx improves accuracy by up to 2.3\% on ETTm1 and remains competitive across four benchmarks.}
  \label{tab:modex_vs_dense_horizon}
\end{table}

\section{Conclusion}
In this study, we explore behaviors of each expert in backbone layer by examining how intermediate features respond to input-sequence dynamics. We introduce layer sensitivity, a gradient-based metric that aggregates element-wise absolute gradients to capture both positive and negative influences. Our analysis reveals that layers at different depths exhibit varying sensitivity to the input sequence’s intrinsic periodicity and trend dynamics, inspiring the construction of MoDEx---Mixture of Depth-specific Experts---composed of lightweight MLPs. MoDEx outperforms existing methods, ranking first on 78\% of 14 results across seven benchmarks, all while requiring significantly fewer parameters and lower computational cost. Furthermore, it serves as a drop-in replacement for self-attention in Transformer variants, consistently boosting accuracy while maintaining its generalizability and efficiency.

\bibliography{aaai2026}

@article{iTransformer,
  author = {Liu, Yong and Hu, Tengge and Zhang, Haoran and Wu, Haixu and Wang, Shiyu and Ma, Lintao and Long, Mingsheng},
  title = {iTransformer: Inverted Transformers are Effective for Time Series Forecasting},
  journal = {International Conference on Learning Representations (ICLR)},
  year = {2024},
  note = {Accepted as a conference paper at ICLR 2024},
  eprint = {2310.06625},
  archivePrefix = {arXiv},
  primaryClass = {cs.LG},
  URL = {https://arxiv.org/abs/2310.06625}
}

@article{vanillaTransformer,
  title={Attention is all you need},
  author={Vaswani, Ashish and Shazeer, Noam and Parmar, Niki and Uszkoreit, Jakob and Jones, Llion and Gomez, Aidan N and Kaiser, {\L}ukasz and Polosukhin, Illia},
  journal={Advances in neural information processing systems},
  volume={30},
  year={2017}
}

@inproceedings{zhou2021informer,
  title={Informer: Beyond efficient transformer for long sequence time-series forecasting},
  author={Zhou, Haoyi and Zhang, Shanghang and Peng, Jieqi and Zhang, Shuai and Li, Jianxin and Xiong, Hui and Zhang, Wancai},
  booktitle={Proceedings of the AAAI conference on artificial intelligence},
  volume={35},
  number={12},
  pages={11106--11115},
  year={2021}
}

@inproceedings{
zhang2023crossformer,
title={Crossformer: Transformer Utilizing Cross-Dimension Dependency for Multivariate Time Series Forecasting},
author={Yunhao Zhang and Junchi Yan},
booktitle={The Eleventh International Conference on Learning Representations },
year={2023},
url={https://openreview.net/forum?id=vSVLM2j9eie}
}

@inproceedings{
liu2022pyraformer,
title={Pyraformer: Low-Complexity Pyramidal Attention for Long-Range Time Series Modeling and Forecasting},
author={Shizhan Liu and Hang Yu and Cong Liao and Jianguo Li and Weiyao Lin and Alex X. Liu and Schahram Dustdar},
booktitle={International Conference on Learning Representations},
year={2022},
url={https://openreview.net/forum?id=0EXmFzUn5I}
}

@inproceedings{
patchtst,
title={A Time Series is Worth 64 Words:  Long-term Forecasting with Transformers},
author={Yuqi Nie and Nam H Nguyen and Phanwadee Sinthong and Jayant Kalagnanam},
booktitle={The Eleventh International Conference on Learning Representations },
year={2023},
url={https://openreview.net/forum?id=Jbdc0vTOcol}
}

@inproceedings{
wang2023micn,
title={{MICN}: Multi-scale Local and Global Context Modeling for Long-term Series Forecasting},
author={Huiqiang Wang and Jian Peng and Feihu Huang and Jince Wang and Junhui Chen and Yifei Xiao},
booktitle={The Eleventh International Conference on Learning Representations },
year={2023},
url={https://openreview.net/forum?id=zt53IDUR1U}
}

@inproceedings{
liu2022scinet,
title={{SCIN}et: Time Series Modeling and Forecasting with Sample Convolution and Interaction},
author={Minhao Liu and Ailing Zeng and Muxi Chen and Zhijian Xu and Qiuxia LAI and Lingna Ma and Qiang Xu},
booktitle={Advances in Neural Information Processing Systems},
editor={Alice H. Oh and Alekh Agarwal and Danielle Belgrave and Kyunghyun Cho},
year={2022},
url={https://openreview.net/forum?id=AyajSjTAzmg}
}

@inproceedings{
wu2023timesnet,
title={TimesNet: Temporal 2D-Variation Modeling for General Time Series Analysis},
author={Haixu Wu and Tengge Hu and Yong Liu and Hang Zhou and Jianmin Wang and Mingsheng Long},
booktitle={The Eleventh International Conference on Learning Representations },
year={2023},
url={https://openreview.net/forum?id=ju_Uqw384Oq}
}

@inproceedings{luo2024moderntcn,
  author = {Luo, Donghao and Wang, Xue},
  title = {ModernTCN: A Modern Pure Convolution Structure for General Time Series Analysis},
  booktitle = {Proceedings of the Twelfth International Conference on Learning Representations (ICLR)},
  year = {2024},
  note = {Published at ICLR 2024},
  url = {https://openreview.net/forum?id=vpJMJerXHU}
}

@article{lee2025casa,
  title={CASA: CNN Autoencoder-based Score Attention for Efficient Multivariate Long-term Time-series Forecasting},
  author={Lee, Minhyuk and Yoon, HyeKyung and Kang, MyungJoo},
  journal={arXiv preprint arXiv:2505.02011},
  year={2025}
}

@inproceedings{dlinear,
  title={Are transformers effective for time series forecasting?},
  author={Zeng, Ailing and Chen, Muxi and Zhang, Lei and Xu, Qiang},
  booktitle={Proceedings of the AAAI conference on artificial intelligence},
  volume={37},
  number={9},
  pages={11121--11128},
  year={2023}
}

@inproceedings{SOFTS,
 author = {Han, Lu and Chen, Xu-Yang and Ye, Han-Jia and Zhan, De-Chuan},
 booktitle = {Advances in Neural Information Processing Systems},
 editor = {A. Globerson and L. Mackey and D. Belgrave and A. Fan and U. Paquet and J. Tomczak and C. Zhang},
 pages = {64145--64175},
 publisher = {Curran Associates, Inc.},
 title = {SOFTS: Efficient Multivariate Time Series Forecasting with Series-Core Fusion},
 url = {https://proceedings.neurips.cc/paper_files/paper/2024/file/754612bde73a8b65ad8743f1f6d8ddf6-Paper-Conference.pdf},
 volume = {37},
 year = {2024}
}

@article{wang2024timemixer,
  title={Timemixer: Decomposable multiscale mixing for time series forecasting},
  author={Wang, Shiyu and Wu, Haixu and Shi, Xiaoming and Hu, Tengge and Luo, Huakun and Ma, Lintao and Zhang, James Y and Zhou, Jun},
  booktitle = {Proceedings of the Twelfth International Conference on Learning Representations (ICLR)},
  year = {2024},
  note = {Published at ICLR 2024},
  url = {https://openreview.net/forum?id=7oLshfEIC2}
}

@inproceedings{ekambaram2023tsmixer,
  title={Tsmixer: Lightweight mlp-mixer model for multivariate time series forecasting},
  author={Ekambaram, Vijay and Jati, Arindam and Nguyen, Nam and Sinthong, Phanwadee and Kalagnanam, Jayant},
  booktitle={Proceedings of the 29th ACM SIGKDD conference on knowledge discovery and data mining},
  pages={459--469},
  year={2023}
}

@inproceedings{
kim2022reversible,
title={Reversible Instance Normalization for Accurate Time-Series Forecasting against Distribution Shift},
author={Taesung Kim and Jinhee Kim and Yunwon Tae and Cheonbok Park and Jang-Ho Choi and Jaegul Choo},
booktitle={International Conference on Learning Representations},
year={2022},
url={https://openreview.net/forum?id=cGDAkQo1C0p}
}

@inproceedings{lai2018modeling,
  title={Modeling long-and short-term temporal patterns with deep neural networks},
  author={Lai, Guokun and Chang, Wei-Cheng and Yang, Yiming and Liu, Hanxiao},
  booktitle={The 41st international ACM SIGIR conference on research \& development in information retrieval},
  pages={95--104},
  year={2018},
}

@inproceedings{ji2023spatio,
  title={Spatio-temporal self-supervised learning for traffic flow prediction},
  author={Ji, Jiahao and Wang, Jingyuan and Huang, Chao and Wu, Junjie and Xu, Boren and Wu, Zhenhe and Zhang, Junbo and Zheng, Yu},
  booktitle={Proceedings of the AAAI conference on artificial intelligence},
  volume={37},
  number={4},
  pages={4356--4364},
  year={2023}
}

@article{mcinnes2018umap,
  title={UMAP: Uniform Manifold Approximation and Projection for Dimension Reduction},
  author={McInnes, Leland and Healy, John and Melville, James},
  journal={stat},
  volume={1050},
  pages={6},
  year={2018}
}

@article{zeevi1996time,
  title={Time series prediction using mixtures of experts},
  author={Zeevi, Assaf and Meir, Ron and Adler, Robert},
  journal={Advances in neural information processing systems},
  volume={9},
  year={1996}
}

@inproceedings{2024mole, title={Mixture-of-linear-experts for long-term time series forecasting}, author={Ni, Ronghao and Lin, Zinan and Wang, Shuaiqi and Fanti, Giulia}, booktitle={International Conference on Artificial Intelligence and Statistics}, pages={4672--4680}, year={2024}, organization={PMLR}
}

@inproceedings{gradcam,
  title={Grad-cam: Visual explanations from deep networks via gradient-based localization},
  author={Selvaraju, Ramprasaath R and Cogswell, Michael and Das, Abhishek and Vedantam, Ramakrishna and Parikh, Devi and Batra, Dhruv},
  booktitle={Proceedings of the IEEE international conference on computer vision},
  pages={618--626},
  year={2017}
}

@article{luo2016understanding,
  title={Understanding the effective receptive field in deep convolutional neural networks},
  author={Luo, Wenjie and Li, Yujia and Urtasun, Raquel and Zemel, Richard},
  journal={Advances in neural information processing systems},
  volume={29},
  year={2016}
}

@inproceedings{ding2022scaling,
  title={Scaling up your kernels to 31x31: Revisiting large kernel design in cnns},
  author={Ding, Xiaohan and Zhang, Xiangyu and Han, Jungong and Ding, Guiguang},
  booktitle={Proceedings of the IEEE/CVF conference on computer vision and pattern recognition},
  pages={11963--11975},
  year={2022}
}

@inproceedings{huang2017densely,
  title={Densely connected convolutional networks},
  author={Huang, Gao and Liu, Zhuang and Van Der Maaten, Laurens and Weinberger, Kilian Q},
  booktitle={Proceedings of the IEEE conference on computer vision and pattern recognition},
  pages={4700--4708},
  year={2017}
}

@article{li2023revisiting,
  title={Revisiting long-term time series forecasting: An investigation on linear mapping},
  author={Li, Zhe and Qi, Shiyi and Li, Yiduo and Xu, Zenglin},
  journal={arXiv preprint arXiv:2305.10721},
  year={2023}
}

\end{document}